\documentclass[runningheads]{llncs}

\usepackage{eccv}

\usepackage{eccvabbrv}

\usepackage{graphicx}
\usepackage{booktabs}

\usepackage[accsupp]{axessibility}  

\usepackage{hyperref}
\usepackage{multirow}
\usepackage{graphicx}
\usepackage{wrapfig}

\usepackage{orcidlink}

\begin{document}

\title{MaDiS: Taming Masked Diffusion Language Models for Sign Language Generation} 

\titlerunning{MaDiS}

\author{Ronglai Zuo \quad Rolandos Alexandros Potamias \quad Qi Sun \quad Evangelos Ververas \\
Jiankang Deng \quad Stefanos Zafeiriou}

\authorrunning{Zuo et al.}

\institute{Imperial College London}

\maketitle
\begin{center}
\centering
\vspace{-3mm}
\includegraphics[width=1.0\textwidth]{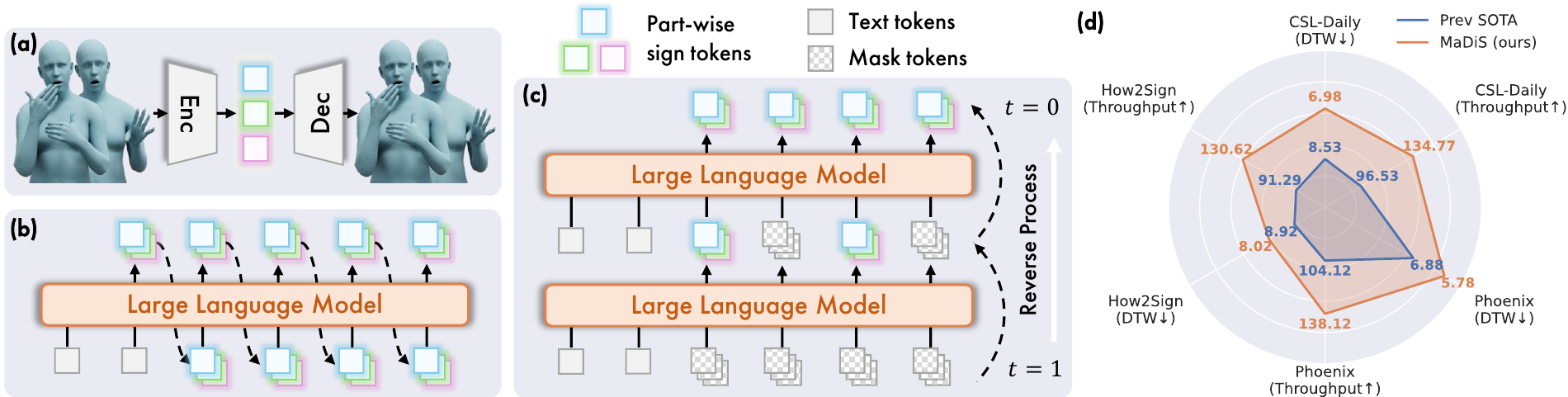}
\vspace{-6mm}
\captionof{figure}{We propose MaDiS, a novel sign language generation approach built upon masked diffusion language models (MDLMs) \cite{llada}. (a) A sign tokenizer discretizes continuous sign motions into part-wise tokens \cite{zuo2025soke}. (b) Conventional autoregressive language models generate tokens in a left-to-right manner, limiting utilization of contexts and inference efficiency. (c) The emerging MDLMs model token distributions with bidirectional contexts and enable parallel multi-token sampling during inference. (d) MaDiS achieves SOTA performance across multiple benchmarks \cite{duarte2021how2sign,csl-daily,2014T} while delivering a 40\% higher throughput.}
\label{fig:teaser}
\vspace{-2mm}
\end{center}

\begin{abstract}
Sign language generation (SLG) aims to translate written texts into expressive sign motions, bridging communication barriers for the Deaf and Hard-of-Hearing communities. Recent studies formulate SLG within the language modeling framework using autoregressive language models, which suffer from unidirectional context modeling and slow token-by-token inference. To address these limitations, we present MaDiS, a masked-diffusion-based language model for SLG that captures bidirectional dependencies and supports efficient parallel multi-token generation. We further introduce a tri-level cross-modal pretraining scheme that jointly learns from token-, latent-, and 3D physical-space objectives to leverage complementary, multi-level sign representations. To accelerate model convergence in the fine-tuning stage, we design a novel unmasking strategy with temporal checkpoints, which restructures generation in a coarse-to-fine manner and reduces the combinatorial complexity of unmasking orders by over $10^{41}$ times. In addition, a mixture-of-parts embedding layer is developed to effectively fuse information stored in different part-wise sign tokens through a learnable gate and well-optimized codebooks. Extensive experiments on CSL-Daily, Phoenix-2014T, and How2Sign demonstrate that MaDiS achieves superior performance across multiple metrics, including DTW error and two newly introduced metrics, SiBLEU and SiCLIP, while delivering a 40\% higher throughput. Code and models will be publicly released.

\vspace{-3mm}
\keywords{Sign Language Generation \and Diffusion Language Model \and Sign Language Pretraining}
\end{abstract}
\vspace{-3mm}
\section{Introduction}
\vspace{-3mm}
\label{sec:intro}

Sign language is the primary communication mode for Deaf and Hard-of-Hearing (DHH) individuals, yet in predominantly oral societies they are often encouraged to rely on text instead of their preferred sign language \cite{yin2021including}. Research in sign language translation (SLT, sign-to-text) and generation (SLG, text-to-sign) aims to improve the accessibility of sign language, thereby mitigating the communication gap between the hearing and DHH communities. While SLT has seen significant progress \cite{li2025uni-sign,gong2024llms,chentwo,jang2025lost,wongsign2gpt,guo2025bridging,fish2025geo-sign,gan2025mixsigngraph,yao2023sign}, SLG remains relatively under-explored.

Early SLG approaches typically rely on glosses, the written form of signs, as an intermediate representation and formulate SLG as a text-to-gloss-to-sign pipeline \cite{zuo2024simple,saunders2022signing,saunders2020progressive,stoll2020text2sign}. While glosses encode rich linguistic information, the performance of gloss-based methods is constrained by the quality of the text-to-gloss module: due to the high cost of gloss annotations, gloss-based datasets are often limited in scale, leading to degraded text-to-gloss performance. For example, Gemini-generated glosses exhibit an error rate of over 60\% \cite{guo2025bridging}. In addition, such cascaded pipelines incur higher inference latency.

Given these limitations, recent efforts have increasingly adopted gloss-free SLG, directly generating signs from text. Typical works \cite{Shi_2025_ICCV,baltatzis2024nsa,fang2023signdiff,wang2025signvip} formulate the task as a visual content generation problem, using either GANs \cite{saunders2022signing} or diffusion models \cite{latentdiff}. 
Recently, motivated by the linguistic character of sign language and the generalizability of large language models (LLMs), a new research direction has emerged that models SLG through tokenization (\cref{fig:teaser}a) and autoregressive language models (ARLMs) \cite{zuo2025soke,yin-etal-2024-t2s,yu2025signavatars,fang2025signllm,fg2024signavatar} (\cref{fig:teaser}b). However, ARLMs have two main limitations: their left-to-right generation order restricts the use of bidirectional contexts and their one-token-at-a-time decoding introduces an inference bottleneck.

To address such problems, we propose MaDiS (\cref{fig:teaser}c), a novel SLG approach built upon an emerging and promising language modeling paradigm, the masked diffusion language model (MDLM) \cite{llada,yang2025mmada,llada-v,smdm}. Unlike conventional ARLMs, MDLMs are based on masked diffusion models \cite{shi2024simplified,sahoo2024simple,ou2025your}, which introduce a forward data masking process and train a mask predictor to approximate the reverse denoising process. This mechanism enables MDLMs to construct model distributions with bidirectional dependencies, allowing them to capture richer contextual information. During the reverse process, multiple tokens can be sampled simultaneously, thereby improving inference efficiency.

Pretraining plays a crucial role in sign language understanding models \cite{li2025uni-sign,hu2023signbert+,zhou2025scaling,Wong_2025_signrep,zhou2023gloss,jiao2024visual}, yet pretraining for SLG models remains largely unexplored. MDLMs are typically pretrained in the token space using a mask-then-predict framework \cite{llada}, where a subset of tokens is randomly masked and subsequently predicted by the model. 
Although such token-space objective is well-established in natural language processing, recent advances in self-supervised learning \cite{jepa,v-jepa,baldassarre2025back} have demonstrated that latent-space pretraining is often more effective in a wide range of vision tasks, including image and video understanding~\cite{mo2024connecting,v-jepa} and world modeling \cite{dino-wm,v-jepa2}.
Moreover, sign language, as a visual language, conveys most of its semantics through body and hand motions in the 3D physical space. Motivated by the advantages of different representation spaces, we propose a tri-level cross-modal pretraining method in which the model learns to predict masked text and sign tokens, latent features, and 3D sign motions simultaneously. This challenging and comprehensive pretraining objective encourages the model to fully exploit complementary sign representations and yields significant performance improvements in the downstream SLG task.

After pretraining, we perform supervised fine-tuning conditioned on unmasked text tokens. The original MDLM can be viewed as an any-order ARLM \cite{seed-diffusion}. Due to its unconstrained confidence-based unmasking strategy, the number of possible unmasking orders is enormous. For example, generating 100 tokens in 25 steps results in approximately $10^{123}$ possible orders. Although this property enables MDLMs to perform well in data-constrained scenarios, it also leads to significantly slower convergence, often requiring more training epochs \cite{prabhudesai2025diffusion}.
To address this issue, we propose a novel unmasking strategy with temporal checkpoints. Inspired by the effectiveness of VAR~\cite{tian2024var}, which generates images progressively from low to high resolution, we insert checkpoints composed of uniformly distributed tokens at noise levels 0.75 and 0.5, corresponding to temporal resolutions of $T/4$ and $T/2$. This design decomposes generation in a coarse-to-fine manner and dramatically prunes unmasking orders by over $10^{41}$ times, thereby accelerating convergence during training.
Furthermore, to better integrate part-wise sign tokens \cite{zuo2025soke}, we propose a plug-and-play mixture-of-parts (MoP) embedding layer. Unlike the previous SOTA method~\cite{zuo2025soke} that simply averages token embeddings from different body parts, our MoP embedding layer is built upon well-optimized VQ-VAE codebooks, and part-wise embeddings are dynamically fused through a learnable gate.
In summary, our contributions are as follows:
\begin{itemize}
    \item We propose MaDiS, a novel gloss-free SLG approach built upon MDLMs that enable bidirectional context modeling and parallel multi-token generation.
    
    \item A tri-level cross-modal pretraining framework is developed that jointly learns from token, latent, and 3D physical spaces, enabling the model to exploit the benefits from complementary, multi-level sign representations.
    
    \item To accelerate training convergence, we design a novel token unmasking strategy with temporal checkpoints, which significantly reduces unmasking order complexity through a coarse-to-fine decomposition. To enhance the integration of part-wise sign tokens, we introduce an MoP embedding layer that is conditioned on VQ-VAE codebooks and controlled by a learnable gate.
    
    \item Extensive experiments on widely adopted benchmarks \cite{duarte2021how2sign,csl-daily,2014T} demonstrate that MaDiS achieves SOTA performance across multiple metrics, including DTW error and two newly introduced metrics, SiBLEU and SiCLIP, while enhancing inference throughput by 40\% (\cref{fig:teaser}d).
\end{itemize}

\vspace{-4mm}
\section{Related Work}
\vspace{-3mm}

\noindent\textbf{Sign Language Generation (SLG).}
SLG aims to translate texts into sign language outputs represented in various forms, such as skeletons \cite{saunders2021mixed,tang2025discrete,arkushin2023ham2pose,xie2024g2p,tang2022gloss}, RGB videos \cite{wang2025signvip,saunders2022signing,signgen,fang2025stable,fang2023signdiff}, and 3D motions \cite{zuo2025soke,zuo2024simple,yin-etal-2024-t2s,yu2025signavatars,baltatzis2024nsa,bensabath2025text}.
Although real-person videos are more realistic, developing end-to-end sign video generation models is challenging due to the high dimensionality of RGB video data. Therefore, we focus on generating sign motions, which can serve as an intermediate representation for subsequent sign video synthesis \cite{Shi_2025_ICCV,shi2024pose,wang2025signaligner}.
Motivated by the linguistic nature of sign language, recent studies \cite{yu2025signavatars,fg2024signavatar,zuo2025soke,yin-etal-2024-t2s} formulate SLG within a language modeling framework. However, these approaches rely on ARLMs, which are inherently limited by their left-to-right generation process.
In this work, we explore emerging MDLMs for SLG, which model token distributions bidirectionally and enable efficient parallel generation.

\noindent\textbf{Masked Diffusion Language Model (MDLM).}
Current LLMs are mostly built upon ARLMs. Although they demonstrate strong capability, their left-to-right generation order limits both efficiency and context modeling. Recently, MDLMs \cite{llada,bie2025llada2,smdm,gongscaling} have emerged as a new language modeling paradigm, showing promising capability, scalability, and efficiency. MDLMs have also been successfully applied to diverse domains, including multimodal understanding \cite{yang2025mmada}, protein design \cite{wang2024diffusion}, and robotics \cite{wen2025llada-vla}.
A related line of work in human motion generation \cite{guo2024momask,guo2025snapmogen,pinyoanuntapong2025maskcontrol,xia2026signmask} adopts a similar masked generation approach \cite{chang2022maskgit}, but their training objectives are determined heuristically and lack the rigorous theoretical foundation of MDLMs.
In this work, we present the first effort to introduce MDLMs into SLG and propose a temporal-checkpoint unmasking strategy to improve training efficiency.

\begin{figure}[t]
\centering
\includegraphics[width=1.0\linewidth]{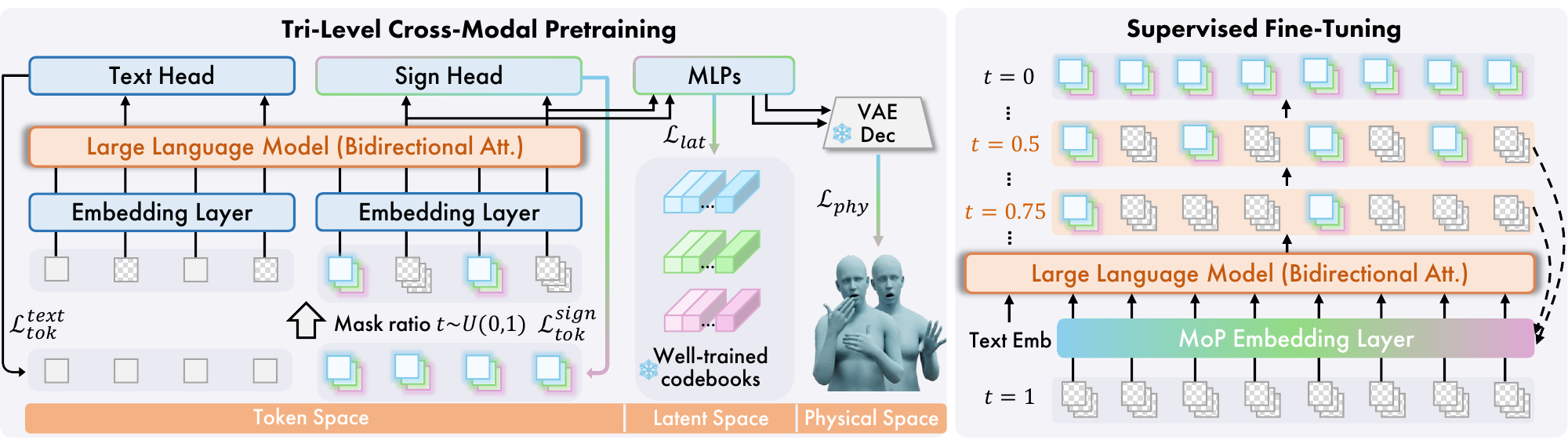}
\vspace{-7mm}
\caption{MaDiS is built upon the emerging masked diffusion language model (MDLM), implemented by modifying a standard decoder-only LLM with bidirectional attention. The MDLM is first pretrained with three objectives ($\mathcal{L}_{tok}^{text} + \mathcal{L}_{tok}^{sign}$, $\mathcal{L}_{lat}$, $\mathcal{L}_{phy}$) from the token, latent, and 3D physical spaces, respectively. We then fine-tune the model conditioned on text inputs using the proposed temporal-checkpoint unmasking strategy and a dedicated mixture-of-parts sign embedding layer.}
\vspace{-5mm}
\label{fig:framework}
\end{figure}

\noindent\textbf{Pretraining for Sign Language Models.}
Pretraining plays a crucial role in sign language understanding tasks. Existing works can be categorized into alignment-based methods \cite{jiao2024visual,zhou2023gloss,kim2025leveraging,ye2024improving,chen2025c2rl}, which pretrain by aligning signs and texts; reconstruction-based methods \cite{hu2023signbert+,Wong_2025_signrep,gueuwou2025shubert,rust-etal-2024-privacy,best,zhao2024masa}, which pretrain by reconstructing sign inputs; hybrid methods \cite{zhou2025scaling}; and next-token prediction methods \cite{li2025uni-sign,zhang2024scaling}.
However, an important oversight is that all these works focus solely on understanding tasks. Our work represents the first attempt to pretrain sign language generation models through a cross-modal and multi-level strategy that jointly predicts sign and text tokens, latent features, and 3D sign motions, thereby effectively injecting sign-aware knowledge into MDLMs.

\vspace{-3mm}
\section{Methodology}
\vspace{-2mm}
An overview of MaDiS is illustrated in \cref{fig:framework}. In the first stage, both text and sign tokens are randomly masked according to a masking ratio $t$, and the MDLM is pretrained with objectives defined across multiple representation spaces (token, latent, and physical) to exploit the benefits from multi-level sign representations. In the subsequent fine-tuning stage, the MDLM keeps text tokens unmasked and learns to generate sign tokens using the proposed unmasking with temporal checkpoints (UTC) strategy and the mixture-of-parts (MoP) embedding layer.

\vspace{-2mm}
\subsection{Preliminaries}
\noindent\textbf{Sign Tokenizer.}
We adopt the open-sourced SOTA decoupled tokenizer \cite{zuo2025soke} to discretize sign motions, which can be reconstructed with high accuracy. It consists of three parallel VQ-VAEs \cite{vqvae} responsible for different body parts. Each VQ-VAE includes an encoder $\mathcal{E}_p$, a decoder $\mathcal{D}_p$, and a codebook $\mathcal{C}_p = \{{\mathbf{c}_p^i}\}_{i=1}^{N_c}$, where $p \in \{b, l, r\}$ represents the upper body, left hand, and right hand, respectively, and $N_c$ is the codebook size. Using this tokenizer, a $T$-frame sign motion sequence $\mathbf{S} \in \mathbb{R}^{T\times d_s}$, where $d_s=133$ denotes the number of SMPL-X parameters \cite{smplx}, is decomposed as three part-wise token sequences $x_p = \{x_p^i\}_{i=1}^{L_s}$, where $x_p^i \in \{1, \dots, N_c\}$ and $L_s$ denotes the token sequence length.

\noindent\textbf{Masked Diffusion Language Model.}
MDLMs \cite{llada,smdm} introduce a new language modeling paradigm based on forward-reverse discrete diffusion. In the forward process, the original token sequence $x_0$ is corrupted with mask tokens $|\mathrm{M}|$ according to the time (noise level) $t \in [0,1]$. The conditional distribution of the masked sequence $x_t$ is defined as $q_{t|0}(x_t|x_0) = \prod_{i=1}^L q_{t|0}(x_t^i|x_0^i)$, where
\vspace{-1.5mm}
\begin{equation}
    q_{t|0}\!\left(x_t^i \mid x_0^i\right) =
    \begin{cases}
    1 - t, & x_t^i = x_0^i,\\[4pt]
    t, & x_t^i = |\mathrm{M}|.
    \end{cases}
\label{eq:forward}
\end{equation}
\vspace{-2.5mm}

The reverse process aims to generate meaningful tokens from the all-masked sequence $x_1$. The conditional distribution is defined as $q_{s|t}(x_s|x_t) = \prod_{i=1}^L q_{s|t}(x_s^i|x_t)$, where $0 \leq s < t \leq 1$, and
\vspace{-2.5mm}
\begin{equation}
    q_{s|t} (x_s^i | x_t) =
    \begin{cases}
    1, &\!\!\!\! x_s^i \! = \! x_t^i \! \neq \! |\mathrm{M}|,\\
    \frac{s}{t}, &\!\!\!\! x_s^i \! = \! x_t^i \! = \! |\mathrm{M}|,\\
    \frac{t-s}{t} q_{0|t} (x_s^i | x_t), &\!\!\!\! x_s^i \! \neq \! |\mathrm{M}|, x_t^i \! = \! |\mathrm{M}|,\\
    0, &\!\!\!\! \text{otherwise.}
    \end{cases}
\label{eq:reverse}
\end{equation}
\vspace{-2.5mm}

A rigorous proof in \cite{ou2025your} shows that $q_{0|t}(x_s^i | x_t) = p_{\text{data}}(x_0^i | x_t^{\mathrm{UM}})$, where $x_t^{\mathrm{UM}}$ denotes the unmasked tokens in $x_t$ that are identical to those in $x_0$. This implies that the time condition $t$ can be omitted from the model inputs, leading to a simplified implementation of MDLM in which the model only needs to predict masked tokens conditioned on the unmasked ones.

\vspace{-3mm}
\subsection{Tri-Level Cross-Modal Pretraining}
Existing works \cite{chang2024large,tirumala2022memorization} reveal that the knowledge of LMs mainly comes from pretraining. As shown in \cref{fig:framework}, we pretrain the MDLM at three spaces, enabling it to absorb knowledge from sign representations of different levels.

\noindent\textbf{Token Space.}
At the first level, we pretrain the MDLM by predicting masked tokens. Following \cite{llada}, at each training iteration, we uniformly sample a noise level (mask ratio) $t \sim U(0,1)$. Denoting text tokens as $x_e = \{x_e^i\}_{i=1}^{L_e}$ and an end-of-sentence token as $|\mathrm{eos}|$, the input sequence is formed by concatenating text tokens, $|\mathrm{eos}|$, and sign tokens: $x_0 = \mathrm{cat}(x_e, |\mathrm{eos}|, x_{\{p\}}, |\mathrm{eos}|) = \{x_0^i\}_{i=1}^{L_e+L_s+2}$. For simplicity, we use $\{p\}$ to denote all part-wise sign tokens, since they are masked and unmasked simultaneously once their index is selected. 
Each index $i$ is selected with a probability of $t$, and its corresponding token is masked to obtain $x_t$.
The token-space loss is defined as a cross-entropy loss \cite{llada}:
\vspace{-2mm}
\begin{equation}
\label{eq:loss_tok}
    \mathcal{L}_{tok} = -\mathbb{E}_{t, x_0, x_t} \left[ \frac{1}{t} \sum_{i\in I_m} \log p_{\theta}(x_0^i \mid x_t) \right], 
\vspace{-3mm}
\end{equation}
where $I_m$ denote the set of masked indices, and $p_\theta(\cdot)$ denotes the token probabilities predicted by the model. 

\noindent\textbf{Latent Space.}
Inspired by the effectiveness of joint-embedding predictive architectures (JEPAs) \cite{jepa,i-jepa} in numerous vision tasks \cite{v-jepa,v-jepa2,dino-wm}, we extend this idea to MDLM pretraining. Unlike the original JEPAs that train an additional target encoder to extract embeddings, we directly leverage the well-trained codebooks $\mathcal{C}_p$, whose embeddings can accurately reconstruct sign motions via the decoders, as the prediction targets to simplify model design.

More specifically, let the last hidden states of the MDLM corresponding to masked sign tokens be $\mathbf{H} \in \mathbb{R}^{|I_s| \times d}$, where $I_s$ denotes the index set of masked sign tokens. We use three MLPs to map $\mathbf{H}$ into the spaces of the codebooks: $\mathbf{H}_p = \mathrm{MLP}_p(\mathbf{H}) \in \mathbb{R}^{|I_s| \times d_c}$, where $d_c$ denotes the dimension of each code embedding. The target embeddings are then collected from the corresponding code embeddings: $\mathbf{\hat{H}}_p = \mathrm{cat}(\{ \mathbf{c}^{n_i}_p \mid i \in I_s, 1 \leq n_i \leq N_c \})$, where $n_i$ denotes the code index of the $i$-th token. Finally, the MDLM learns to predict the code embeddings with a smoothed L1 loss ($\|\cdot\|_1$): $\mathcal{L}_{lat} = \sum_{p} \| \mathbf{H}_p - \mathbf{\hat{H}}_p \|_1$.

\noindent\textbf{Physical Space.}
Inspired by the visual-language nature of sign languages, we incorporate 3D sign motion reconstruction as an additional pretraining objective in the physical space. More specifically, we reuse the well-trained VAE decoder $\mathcal{D}_p$, which is frozen and used to reconstruct sign motions from the mapped hidden states $\mathbf{H}_p$. Denoting the ground-truth part-wise sign motions at the masked indices as $\mathbf{\hat{S}}_p$, the objective function in the physical space is defined as: $\mathcal{L}_{phy} = \sum_{p} \| \mathcal{D}_p(\mathbf{H}_p) - \mathbf{\hat{S}}_p \|_1$.

Unlike existing works \cite{hu2023signbert+,zhou2025scaling,rust-etal-2024-privacy} that mask and reconstruct sign poses, we mask tokens but reconstruct poses (motions), which is more challenging and encourages the model to learn more grounded sign representations. The overall pretraining loss is defined as $\mathcal{L}_{pre} = \mathcal{L}_{tok} + \mathcal{L}_{lat} + \mathcal{L}_{phy}$.

\vspace{-3mm}
\subsection{Supervised Fine-Tuning}
\vspace{-1mm}
\begin{figure}[t]
\centering
\includegraphics[width=0.95\linewidth]{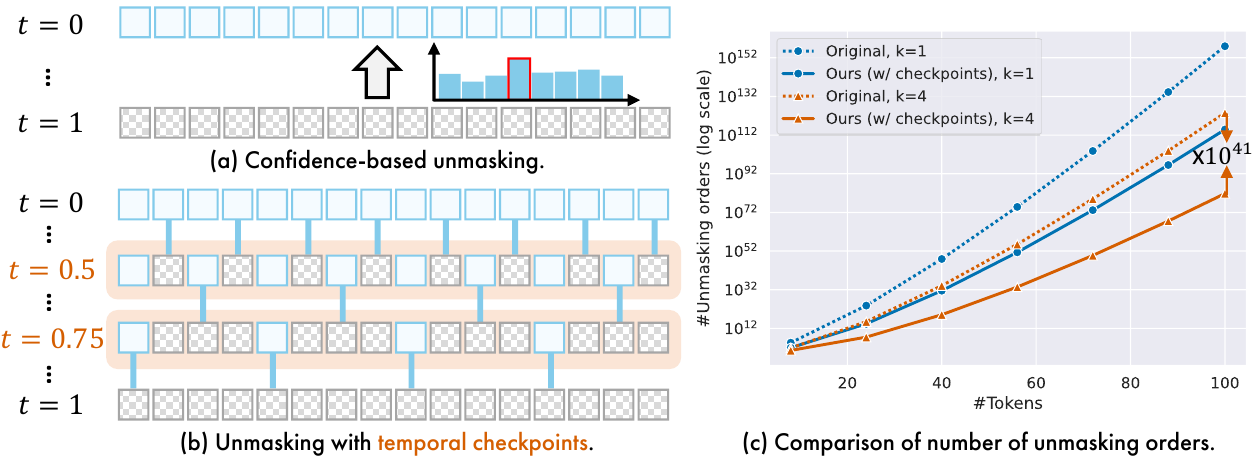}
\vspace{-3mm}
\captionof{figure}{(a) Vanilla MDLMs adopt an unconstrained confidence-based unmasking strategy \cite{llada}. (b) We insert temporal checkpoints at noise levels $0.75$ and $0.5$ to reduce the complexity of unmasking orders. (c) The original unmasking strategy allows arbitrary unmasking orders (about $10^{123}$ for generating 100 tokens), while our method reduces this complexity by over $10^{41}$ times.}
\vspace{-5mm}
\label{fig:unmasking}
\end{figure}

After the pretraining stage, the MDLM is expected to acquire basic sign language knowledge and adapt to the diffusion process with bidirectional contexts. We then fine-tune the entire model while keeping the text tokens unmasked.

\noindent\textbf{Unmasking with Temporal Checkpoints (UTC).}
During the reverse process, the original MDLM \cite{llada} adopts a simple confidence-based unmasking strategy (\cref{fig:unmasking}a). Specifically, at each step, the model unmasks $k$ tokens with the highest confidence, defined as the maximum of their categorical probabilities. However, this unconstrained strategy leads to highly diverse unmasking orders. Denoting the number of generated tokens as $M$, the total number of possible unmasking orders is:
\vspace{-2mm}
\begin{equation}
\label{eq:conf}
    N_u = \binom{M}{k}\binom{M-k}{k}\cdots\binom{k}{k} = \frac{M!}{(k!)^{M/k}}.
\vspace{-2mm}
\end{equation}
When $M=100$ and $k=4$, this number can achieve $2.9\times10^{123}$. Although such diversity can enhance model performance, it also results in slow convergence \cite{prabhudesai2025diffusion}.

Inspired by VAR \cite{tian2024var}, which adopts a coarse-to-fine generation scheme with progressively enhanced resolution, we propose a novel unmasking strategy with temporal checkpoints (UTC), as shown in \cref{fig:unmasking}b. Specifically, we divide the reverse process into three stages at $t=0.75$ and $t=0.5$, where $1/4$ and $1/2$ of the tokens have been unmasked, respectively. Both pivot states are enforced to contain uniformly distributed tokens, while a standard confidence-based unmasking strategy is applied within each stage. Consequently, the number of possible orders under our UTC is substantially reduced to:
\vspace{-1mm}
\begin{equation}
\label{eq:utc}
N_u^{\mathrm{UTC}} = \frac{(M/4)!}{(k!)^{M/4k}} \cdot \frac{(M/4)!}{(k!)^{M/4k}} \cdot \frac{(M/2)!}{(k!)^{M/2k}}.
\vspace{-1mm}
\end{equation}
When $M=100$ and $k=4$, UTC reduces the number of orders by $10^{41}\times$ compared to vanilla unmasking strategy (\cref{fig:unmasking}c). Moreover, the resulting generation process follows a coarse-to-fine structure, with both factors jointly easing model learning and accelerating convergence.
To ensure training-inference consistency, we employ predefined index masks to filter out infeasible states during training. More implementation details are provided in the supplement.

\begin{wrapfigure}{r}{0.5\linewidth}
    \centering
    \vspace{-9mm}
    \includegraphics[width=\linewidth]{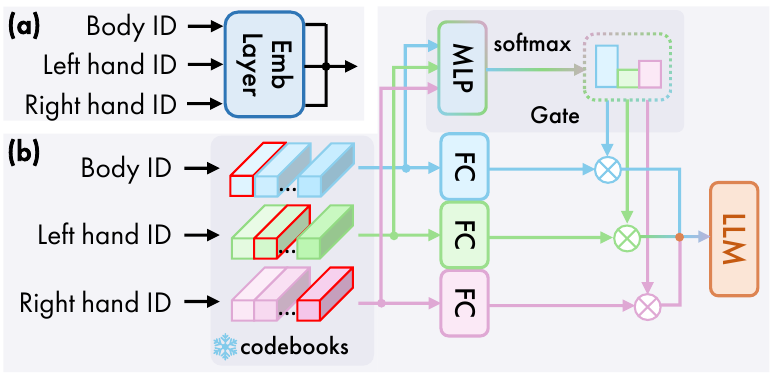}
    \vspace{-8mm}
    \caption{(a) A typical sign embedding layer averages part-wise token embeddings \cite{zuo2025soke}. (b) The proposed mixture-of-parts embedding layer leverages optimized VAE codebooks and uses a learnable gate to control contributions from different body parts.}
    \vspace{-6mm}
    \label{fig:sign_emb}
\end{wrapfigure}

\noindent\textbf{Mixture-of-Parts (MoP) Embedding Layer.}
In contrast to the pretraining stage, where different part-wise tokens are treated equally, we aim to more discriminatively capture the relations among body parts during fine-tuning with text conditions. To this end, our MoP embedding layer is built upon the well-trained VAE codebooks and employs a learnable gate to control the contributions from different parts. For instance, since numerous signs are produced with a single hand \cite{battison1978lexical,forte2023reconstructing}, it would be beneficial to assign distinct weights to each hand.

As shown in \cref{fig:sign_emb}b, given the part IDs ($i_p$), the part-wise code embeddings are passed through fully-connected (FC) layers to match the model dimension of MDLM: $\mathbf{E}_p = \mathrm{FC}_p(\mathbf{c}_p^{i_p})$. Furthermore, we use an MLP with a softmax layer to learn gating weights: $\mathbf{G} = \{g_p\} = \mathrm{softmax}(\mathrm{MLP}(\mathrm{cat}(\{\mathbf{c}_p^{i_p}\})))$. The final sign embedding is obtained as $\mathbf{E} = \sum_p g_p \cdot \mathbf{E}_p$.

Similar to \cite{jiao2024visual}, we also observe that continuing to use pretraining objectives during fine-tuning is beneficial. Therefore, the fine-tuning loss is defined as $\mathcal{L}_{sft} = \mathcal{L}_{tok} + \alpha (\mathcal{L}_{lat} + \mathcal{L}_{phy})$, where $\alpha = 0.5$ is determined empirically through a sensitivity analysis in the supplement.

\vspace{-3mm}
\subsection{SiBLEU and SiCLIP}
\vspace{-1mm}
Existing SLG models \cite{zuo2025soke,baltatzis2024nsa,zuo2024simple,arkushin2023ham2pose} are typically evaluated using dynamic time warping (DTW) error \cite{arkushin2023ham2pose} and back-translation (BT) \cite{saunders2020progressive}. However, DTW error may over-optimistically reflect model performance since it measures the minimum distance between generated and ground-truth sequences. Moreover, its computation is inefficient due to the reliance on dynamic programming. 
On the other hand, BT evaluates performance in the text space, which overlooks the visual and multi-cue nature of sign language, and the translation process itself introduces unavoidable errors \cite{SiLVERScore}.
To address these issues, we propose two new metrics that serve as effective alternatives.

\noindent\textbf{SiBLEU -- Straightforward and Efficient Evaluation in the Token Space.}
For LM-based SLG approaches, signs are tokenized, which enables a straightforward evaluation by computing BLEU scores \cite{papineni2002bleu} over generated tokens against ground-truth tokens. We term this the SiBLEU score. A higher SiBLEU indicates greater overlap with the ground-truth tokens.

\noindent\textbf{SiCLIP -- Evaluation in a Joint Embedding Space.}
In related domains such as human motion generation \cite{tevet2022motionclip,meng2025rethinking} and text-to-video generation \cite{zhang2024controlvideo,wu2023tune-a-video,xing2024simda}, CLIP \cite{clip} is widely used to evaluate the alignment between prompts (texts) and generations in a joint-embedding space. Motivated by this, we train a CLIP model over ground-truth sign-text pairs in the training sets, and adopt retrieval-based metrics \cite{bensabath2025text,petrovich2023tmr,jiang2024signclip} for evaluating the semantic alignment between texts and generated signs. More details are provided in the supplement.
\begin{figure*}[t]
    \centering
    \includegraphics[width=1.0\linewidth]{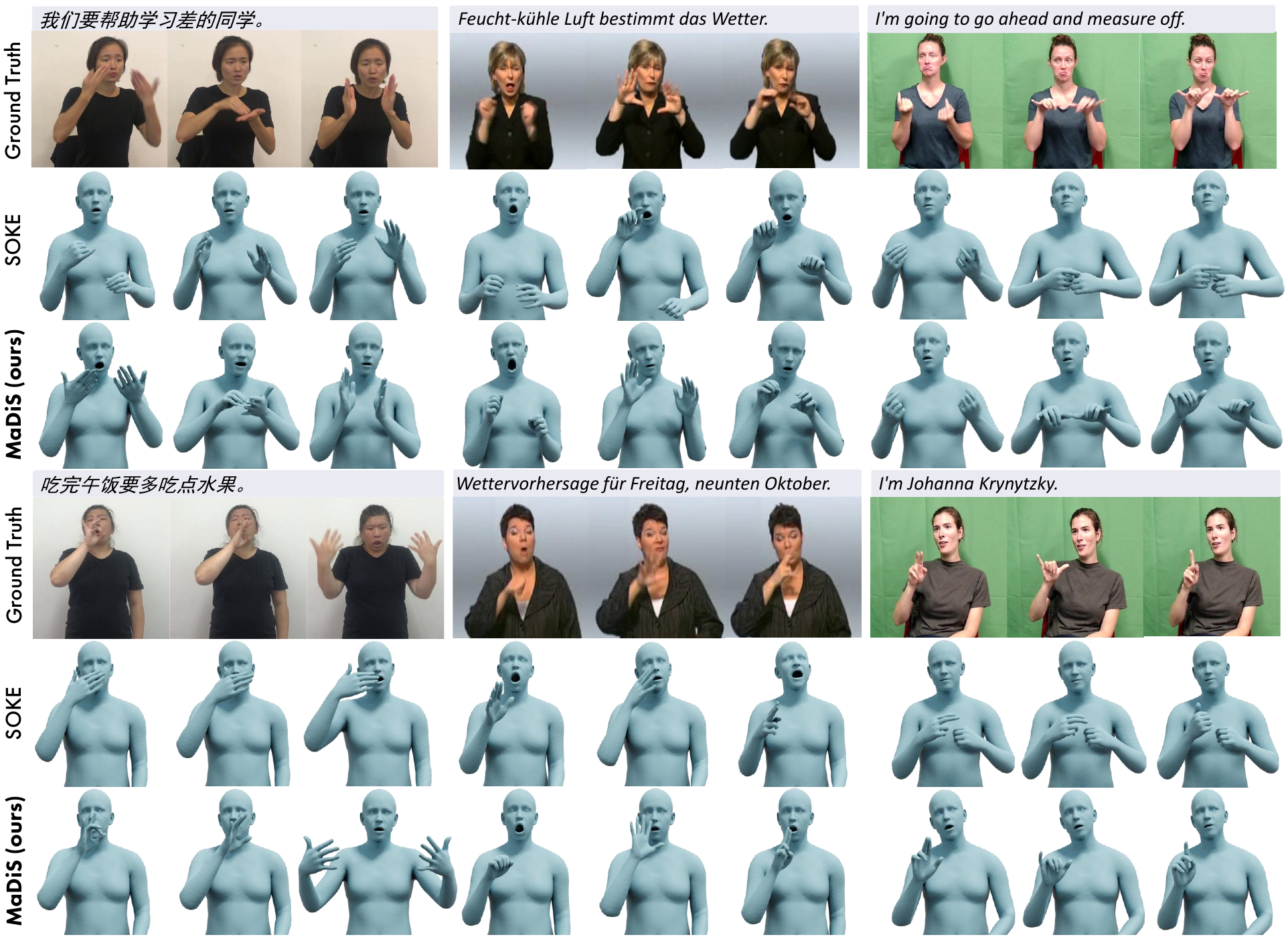}
    \vspace{-7mm}
    \caption{Qualitative comparisons of generated signs between our proposed method, MaDiS, with the SOTA method, SOKE \cite{zuo2025soke}, on the test sets of CSL-Daily (left), Phoenix-2014T (middle), and How2Sign (right).}
    \label{fig:qual}
    \vspace{-5mm}
\end{figure*}

\vspace{-4mm}
\section{Experiments}
\vspace{-2mm}

\noindent\textbf{Datasets.}
We evaluate MaDiS on three widely-adopted datasets: CSL-Daily \cite{csl-daily}, Phoenix-2014T \cite{2014T}, and How2Sign \cite{duarte2021how2sign}, which contain 20K/8K/35K video-text pairs for Chinese, American, and German sign languages, respectively. We use the curated SMPL-X poses provided in \cite{zuo2025soke} as sign motion representations.

\noindent\textbf{Evaluation Metrics.}
Due to the length difference between generated signs and ground truth, we follow the previous work \cite{zuo2025soke} to use dynamic time warping over joint position errors (DTW-JPE) as the major metric to measure sequence-level distances. Note that we do not compute procrustes alignment (PA) since it will align rotation before computing joint errors, leading to overoptimistic performance \cite{black2023bedlam}. As a complement, we use SiBLEU-4 as an indicator for token-level precision and report the sentence-level sign-to-text retrieval performance (R@1) \cite{bensabath2025text,cheng2023cico} using SiCLIP as a joint feature extractor for signs and texts.

\noindent\textbf{Implementation Details.}
We instantiate the MDLM with Qwen3-0.6B-Base \cite{yang2025qwen3}, a leading open-source LLM supporting 119 languages. To enable bidirectional modeling, we replace the causal attention mask in each transformer layer with a non-causal one \cite{llada}. During pretraining, the model is trained on the combined training sets of all three datasets to maximize data utilization \cite{hu2023signbert+,zhang2024scaling}. In the fine-tuning stage, it is trained on each dataset independently. We empirically set $k=4$ and $M=100$, which corresponds to 400 frames after decoding due to the 4x up-sampling layer in the VAE decoder. This length is sufficient to cover about 98\% of training videos, and over-length videos are uniformly sampled to fit within 400 frames. When generating sign tokens, we adopt multi-head decoding \cite{zuo2025soke} to predict all three part-wise tokens simultaneously, and tokens appearing after the end-of-sentence token are discarded \cite{llada}. The initial learning rate is set to $2e-4$, and we adopt the AdamW optimizer \cite{adamw} with a cosine learning rate scheduler and a batch size of 64 per GPU for both stages. The models are trained for 200 and 150 epochs in the pretraining and fine-tuning stages, respectively, using 4x Nvidia GH200 GPUs.

\begin{table*}[t]
\small
\setlength\tabcolsep{2pt}
    \centering
    \resizebox{\linewidth}{!}{
    \begin{tabular}{l|ccccc|ccccc|ccccc}
    \toprule
    \multirow{3}{*}{Method} & \multicolumn{5}{c|}{CSL-Daily} & \multicolumn{5}{c|}{Phoenix-2014T} & \multicolumn{5}{c}{How2Sign} \\
    & \multicolumn{2}{c}{DTW-JPE$\downarrow$} & \multicolumn{2}{c}{SiBLEU$\uparrow$} & SiCLIP$\uparrow$ & \multicolumn{2}{c}{DTW-JPE$\downarrow$} & \multicolumn{2}{c}{SiBLEU$\uparrow$} & SiCLIP$\uparrow$ & \multicolumn{2}{c}{DTW-JPE$\downarrow$} & \multicolumn{2}{c}{SiBLEU$\uparrow$} & SiCLIP$\uparrow$ \\
    \cmidrule(r){2-3}\cmidrule(r){4-5}\cmidrule(r){6-6}\cmidrule(r){7-8}\cmidrule(r){9-10}\cmidrule(r){11-11}\cmidrule(r){12-13}\cmidrule(r){14-15}\cmidrule(r){16-16}
    & Body & Hands & Body & Hands & R@1 & Body & Hands & Body & Hands & R@1 & Body & Hands & Body & Hands & R@1 \\
    
    \midrule
    Prog. Trans.$^*$ \cite{saunders2020progressive} & 16.30 & 32.63 & \multicolumn{2}{c}{N/A} & 13.65 & 15.01 & 31.77 & \multicolumn{2}{c}{N/A} & 14.03 & 14.74 & 30.17 & \multicolumn{2}{c}{N/A} & 9.88 \\
    Text2Mesh$^*$ \cite{stoll2022there}& 13.76 & 30.37 & \multicolumn{2}{c}{N/A} & 15.84 & 14.04 & 31.64 & \multicolumn{2}{c}{N/A} & 12.08 & 15.50 & 32.97 & \multicolumn{2}{c}{N/A} & 9.30 \\
    T2S-GPT$^*$ \cite{yin-etal-2024-t2s} & 12.32 & 15.43 & 2.11 & N/A & 20.03 & 11.65 & 19.09 & 1.93 & N/A & 22.10 & 12.65 & 18.44 & 2.13 & N/A & 12.58 \\
    NSA \cite{baltatzis2024nsa} & -- & -- & \multicolumn{2}{c}{N/A} & -- & -- & -- & \multicolumn{2}{c}{N/A} & -- & 9.16 & 18.51 & \multicolumn{2}{c}{N/A} & 15.04 \\
    S-MotionGPT$^*$ \cite{jiang2023motiongpt} & 11.58 & 11.31 & 2.26 & N/A & 20.37 & 10.42 & 9.08 & 1.97 & N/A & 32.42 & 12.41 & 13.74 & 2.32 & N/A & 12.77 \\
    MoMask++ \cite{guo2025snapmogen} & 8.93 & 10.57 & 1.53 & N/A & 22.62 & 8.16 & 9.96 & 1.49 & N/A & 37.88 & 9.06 & 11.34 & 1.58 & N/A & 20.39 \\
    SOKE (mBART)$^\dagger$ \cite{zuo2025soke} & 7.38 & 9.68 & 2.59 & 2.14 & 23.64 & 6.04 & 7.72 & 2.71 & 1.85 & 34.30 & 7.75 & 10.08 & 2.52 & 1.74 & 16.12 \\
    SOKE (Qwen3)$^\dagger$ \cite{zuo2025soke} & 7.29 & 9.59 & 2.71 & 2.27 & 25.05 & 6.09 & 7.65 & 2.38 & 1.98 & 38.53 & 7.18 & 10.09 & 2.57 & 1.97 & 18.48 \\
    
    \midrule
    MDLM baseline & 7.12 & 9.57 & 3.17 & 2.55 & 29.17 & 5.98 & 7.58 & 2.36 & 2.14 & 44.25 & 7.02 & 10.16 & 2.73 & 1.79 & 25.78 \\
    MaDiS (ours) & \textbf{5.99} & \textbf{7.96} & \textbf{4.87} & \textbf{4.08} & \textbf{38.61} & \textbf{4.97} & \textbf{6.59} & \textbf{4.54} & \textbf{3.33} & \textbf{57.94} & \textbf{6.59} & \textbf{9.45} & \textbf{4.41} & \textbf{3.46} & \textbf{34.27} \\
    
    \bottomrule
    \end{tabular}
    }
    \caption{Comparison with SOTA sign language generation methods. Note that since SiBLEU scores are computed over sign tokens, we report ``N/A'' for tokenizer-free methods \cite{saunders2020progressive,stoll2022there,baltatzis2024nsa}. Similarly, for methods \cite{yin-etal-2024-t2s,jiang2023motiongpt,guo2025snapmogen} that employ a single tokenizer for whole-body motions, the SiBLEU-Hands results are also marked as ``N/A''. $^*$DTW errors are reported by \cite{zuo2025soke}. $^\dagger$Methods using external sign dictionaries.}
    \label{tab:sota_slg}
    \vspace{-8mm}
\end{table*}

\vspace{-3mm}
\subsection{Comparison with State-of-the-Art Methods}
\vspace{-1mm}
\noindent\textbf{Qualitative Comparison.}
We first conduct qualitative comparisons with the previous SOTA method, SOKE \cite{zuo2025soke}, which proposes a simple autoregressive SLG model without dedicated pretraining and can only capture unidirectional contexts due to the nature of ARLMs. In contrast, our MaDiS is pretrained across multiple sign representation spaces and benefits from an MDLM backbone that enables bidirectional contextual modeling. As shown in \cref{fig:qual}, our model produces more precise and natural sign motions, demonstrating better generalization across diverse hand shapes, palm orientations, single- and double-handed signs, and finger details. Video demos are provided in the supplement.

\noindent\textbf{Quantitative Comparison.}
As detailed in \cref{tab:sota_slg}, we conduct a quantitative comparison with SOTA SLG methods. Since we introduce two new evaluation metrics, we carefully reproduce previous works using their open-source codes \cite{saunders2020progressive,stoll2022there,jiang2023motiongpt,guo2025snapmogen,zuo2025soke} or our own reimplementations \cite{yin-etal-2024-t2s,baltatzis2024nsa}. 
MoMask++ \cite{guo2025snapmogen} is a leading human motion generation method based on a masked generation approach \cite{chang2022maskgit}. However, its masking schedule is heuristically determined without a theoretical foundation, and its multi-scale tokenization introduces error accumulation during inference, leading to inferior performance than SOKE.
For fairness, we also report the performance of SOKE when using Qwen3-0.6B as the LM, instead of the originally adopted mBART-Large \cite{mbart}. Although both LMs have a similar size (0.6B), we observe that employing Qwen3 leads to slightly better performance, which can be attributed to the larger pretraining corpus of the Qwen3 series.

\begin{figure}[t]
    \centering
    \includegraphics[width=1.0\linewidth]{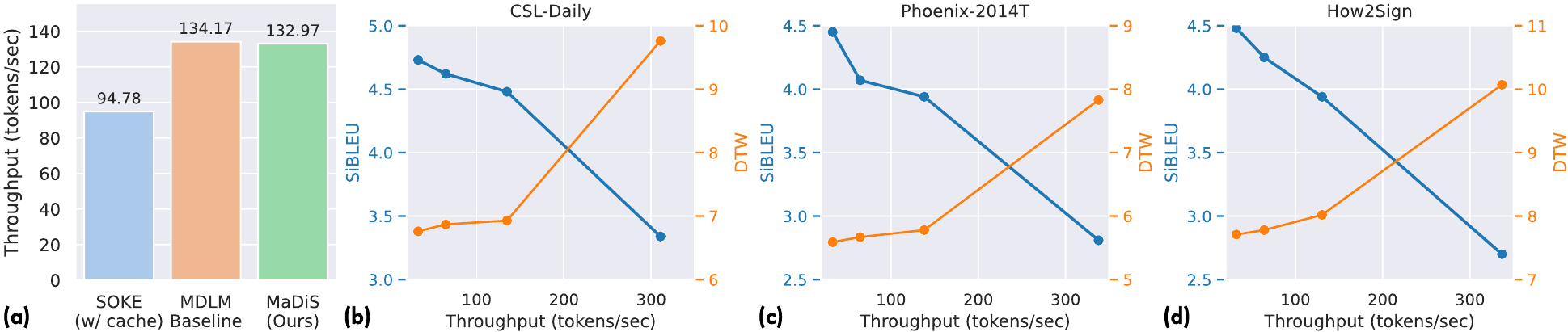}
    \vspace{-7mm}
    \caption{(a) Comparison of throughput measured in output tokens per second. (b-d) Efficiency-quality trade-off on three datasets. The generation length is fixed to 100 tokens, and the number of sampling steps is set to 100, 50, 25 (default), and 10, corresponding to decoding 1, 2, 4, and 10 tokens per step, respectively.}
    \vspace{-4mm}
    \label{fig:throughput}
\end{figure}

Furthermore, we report the performance of an MDLM baseline obtained by directly training an MDLM on sign language datasets, without any of our proposed techniques (pretraining, UTC, or MoP).
Notably, this baseline performs on par with SOKE, which utilizes external sign dictionaries, while offering 40\% higher throughout (\cref{fig:throughput}a). This highlights the potential of MDLMs for SLG.
Finally, our proposed MaDiS achieves new SOTA results across the test sets of all three datasets and all evaluation metrics.

\begin{figure}[t]
    \centering
    \includegraphics[width=1.0\linewidth]{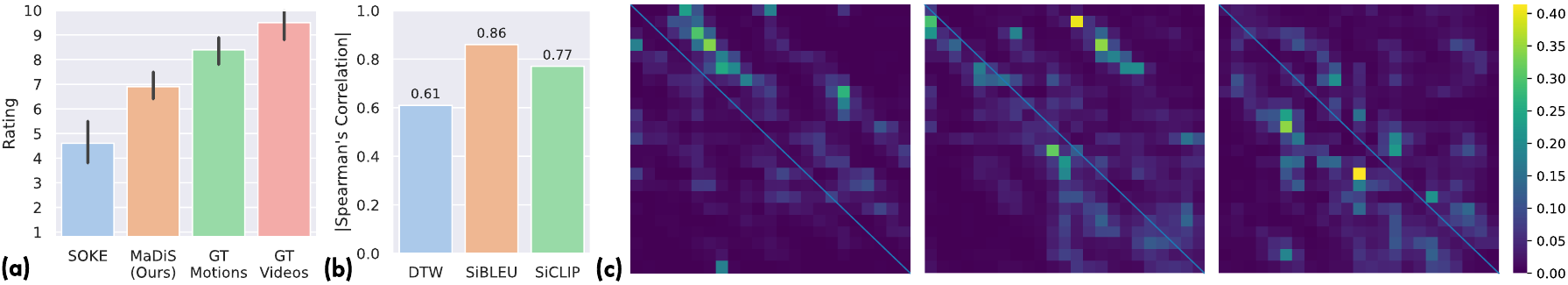}
    \vspace{-7mm}
    \caption{(a) User study with professional CSL signers. We report average ratings of the previous SOTA method, SOKE \cite{zuo2025soke}, our proposed MaDiS, ground-truth motions, and ground-truth RGB videos. Error bars indicate 95\% confidence intervals. (b) Absolute values of Spearman’s rank correlation coefficient \cite{spearman} between signers’ ratings and metric scores. DTW and SiBLEU refer to the average of Body and Hands. (c) Visualization of attention maps from the penultimate Transformer layer.}
    \label{fig:user}
    \vspace{-5mm}
\end{figure}

\noindent\textbf{Efficiency.}
As shown in \cref{fig:throughput}a, across all test samples of the three datasets, our method achieves a throughput of 132 tokens/sec, representing a 40\% improvement over SOKE with KV caching. This high throughput enables the generation of a 400-frame (100-token) video in only 0.76s, thereby satisfying real-time requirements. This improvement stems from the highly efficient parallel token generation enabled by the reverse process in MDLMs.

In \cref{fig:throughput}b-d, we observe that MaDiS provides a flexible trade-off between generation quality and efficiency, where quality is measured by DTW and SiBLEU scores (averaged over Body and Hands). When the number of sampling steps is set to 100, corresponding to sampling one token at a time, the model achieves the highest generation quality but the lowest throughput. In contrast, with only 10 sampling steps, inference becomes extremely fast (exceeding 300 tokens/sec) at the cost of degraded performance. Using 25 sampling steps yields a balance between generation quality and efficiency.

\noindent\textbf{Human Evaluation.}
To assess the intelligibility of the generated signs, we conducted a user study involving five professional CSL signers with 4 to 13 years of signing experience. They were asked to rate the alignment between generated signs and text annotations on a scale from 1 to 10, where higher scores indicate higher sign accuracy and better semantic conveyance. Ground-truth motions and RGB videos were also included for reference. Specifically, we provided 15 generated signs from SOKE and from our proposed MaDiS, together with the corresponding ground-truth motions and videos. The order of all samples was randomly shuffled to avoid potential bias. All participants were compensated above the average hourly rate in China.

As shown in \cref{fig:user}a, our method achieves an average rating of 6.9, substantially outperforming SOKE, which receives an average rating of 4.6. The ground-truth motions obtain an average rating of 8.4, further demonstrating the feasibility of using motions as effective sign language representations.
Furthermore, we report the absolute values of Spearman's rank correlation coefficient \cite{spearman} (1 means high correlation while 0 means no correlation) to study the alignment of the metric scores with human preference (signers' ratings). As shown in \cref{fig:user}b, SiBLEU and SiCLIP exhibit strong correlation, whereas DTW, based on minimum-distance alignment, shows weaker correlation.

\vspace{-3mm}
\subsection{Ablation Studies}
We conduct all ablation studies on CSL-Daily unless otherwise specified. 

\noindent\textbf{Usage of Bidirectional Contexts.}
To verify that the MDLM effectively models bidirectional contexts, we visualize attention maps from the penultimate Transformer layer at the final sampling step, which captures rich contextual interactions while being less influenced by the task-specific output projection of the final layer \cite{clark2019does,tenney2019bert_red}. As shown in \cref{fig:user}c, we select the three attention heads with the highest future-attention ratios, defined as $\sum_{i, j>i} \mathbf{A}_{i,j} / \sum_{i,j} \mathbf{A}_{i,j}$, where $\mathbf{A}$ denotes the attention map, and visualize their patterns. The results show that the model allocates substantial attention to future tokens (upper triangular region), indicating effective bidirectional modeling. Quantitatively, the average future-attention ratio reaches 41\% across all heads over the test set.

\begin{table}[t]
    \small
    \centering
    \begin{minipage}[b]{0.6\linewidth}
        \centering
        \setlength\tabcolsep{2pt}
        \resizebox{\linewidth}{!}{
        \begin{tabular}{cccc|ccccc}
        \toprule
        \multicolumn{2}{c}{Token} & \multirow{2}{*}{Latent} & \multirow{2}{*}{Physical} & \multicolumn{2}{c}{DTW-JPE$\downarrow$} & \multicolumn{2}{c}{SiBLEU$\uparrow$} & \multicolumn{1}{c}{SiCLIP$\uparrow$} \\
        \cmidrule(r){1-2}\cmidrule(r){5-6}\cmidrule(r){7-8}\cmidrule(r){9-9}
        Sign & Text & & & Body & Hands & Body & Hands & R@1 \\
        \midrule
    
        & & & & 6.93 & 9.27 & 3.48 & 2.84 & 31.49 \\
        \checkmark & & & & 6.81 & 8.90 & 3.76 & 2.97 & 33.01 \\
        \checkmark & \checkmark & & & 6.41 & 8.64 & 4.18 & 3.24 & 34.76 \\
        \checkmark & \checkmark & \checkmark & & 6.22 & 8.39 & 4.45 & 3.67 & 35.96 \\
        \checkmark & \checkmark & \checkmark & \checkmark & \textbf{5.99} & \textbf{7.96} & \textbf{4.87} & \textbf{4.08} & \textbf{38.61} \\
        
        \bottomrule
        \end{tabular}
        }
        \caption{Ablation study for the tri-level cross-modal pretraining.}
        \label{tab:abl_pretrain}
    \end{minipage}
    \hfill
    \begin{minipage}[b]{0.39\linewidth}
        \centering
        \includegraphics[width=\linewidth]{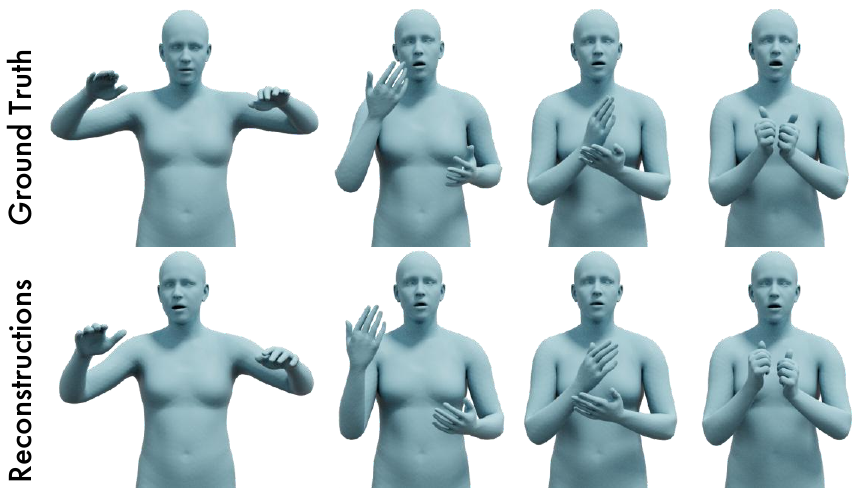}
        \captionof{figure}{Visualization of reconstructed signs from mask tokens.}
        \label{fig:vis_recon}
    \end{minipage}
    \vspace{-12mm}
\end{table}

\noindent\textbf{Tri-Level Cross-Modal Pretraining.}
We pretrain the model across three spaces (token, latent, and physical) to fully exploit the benefits from multi-level sign representations. Results in \cref{tab:abl_pretrain} validate the effectiveness of each pretraining level.
First, at the token level, masking both sign and text tokens yields better performance than masking sign tokens alone, demonstrating the benefit of cross-modal training. This observation is consistent with previous findings on alignment-based pretraining in sign language translation \cite{jiao2024visual,zhou2023gloss,low2025sage}.
Second, predicting latent features proves effective in reducing DTW errors and enhancing sentence-level semantic alignment.
Finally, incorporating physical-space training by reconstructing sign motions leads to the best overall performance. 
This design also enables the pretrained model to learn physically grounded representations, allowing it to accurately reconstruct sign motions from masked tokens (\cref{fig:vis_recon}).
Quantitatively, when $t=0.75$ (\ie, masking 25\% of tokens), the reconstructed signs achieve mean position errors of 24.79/5.64mm on body/hand joints, comparable to the performance of the sign tokenizer \cite{zuo2025soke}.

\begin{table}[t]
    \small
    \centering
    \begin{minipage}[b]{0.6\linewidth}
        \centering
        \setlength\tabcolsep{1pt}
        \resizebox{\linewidth}{!}{
        \begin{tabular}{l|ccccc}
        \toprule
        \multirow{2}{*}{Checkpoint} & \multicolumn{2}{c}{DTW-JPE$\downarrow$} & \multicolumn{2}{c}{SiBLEU$\uparrow$} & \multicolumn{1}{c}{SiCLIP$\uparrow$} \\
        \cmidrule(r){2-3}\cmidrule(r){4-5}\cmidrule(r){6-6}
        (distribution, noise level) & Body & Hands & Body & Hands & R@1 \\
        \midrule
    
        N/A & 6.61 & 9.05 & 3.87 & 3.11 & 32.81 \\
        Uniform, $t=0.75$ & 6.33 & 8.73 & 4.15 & 3.52 & 34.09 \\
        Uniform, $t=0.5$ & 6.18 & 8.23 & 4.63 & 3.94 & 36.73 \\
        Uniform, $t=0.75/0.5$ & \textbf{5.99} & \textbf{7.96} & \textbf{4.87} & \textbf{4.08} & \textbf{38.61} \\
        Contiguous, $t=0.75/0.5$ & 6.28 & 8.44 & 4.31 & 3.80 & 34.85 \\
        
        \bottomrule
        \end{tabular}
        }
        \caption{Ablation study for unmasking with temporal checkpoints.}
        \label{tab:abl_unmask}
    \end{minipage}
    \hfill
    \begin{minipage}[b]{0.39\linewidth}
        \centering
        \includegraphics[width=\linewidth]{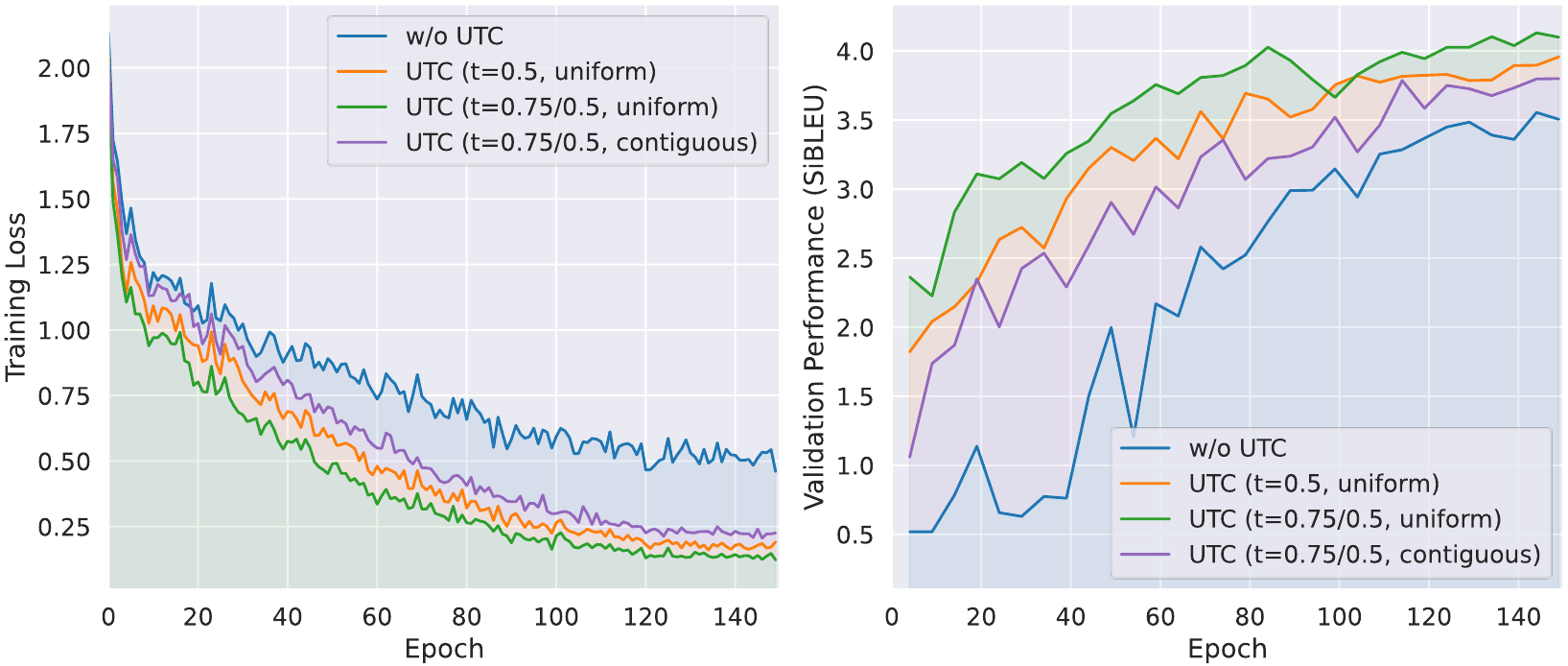}
        \captionof{figure}{Visualization of training loss (left) and validation performance (average of SiBLEU-Body and SiBLEU-Hands, right).}
        \label{fig:utc_curve}
    \end{minipage}
    \vspace{-11mm}
\end{table}

\noindent\textbf{Unmasking with Temporal Checkpoints (UTC).}
To accelerate model convergence, we propose a novel unmasking strategy with temporal checkpoints, which substantially reduces the complexity of unmasking orders in a coarse-to-fine manner. As shown in \cref{tab:abl_unmask}, inserting checkpoints composed of uniformly distributed tokens at either $t=0.75$ ($T/4$ temporal resolution) or $t=0.5$ ($T/2$ temporal resolution) yields superior performance over the no-checkpoint setting. 
We further construct a baseline that retains only the first $1/4$ or $1/2$ contiguous tokens at the pivot states. While this design also reduces the order complexity, it does not explicitly model the coarse-to-fine prior. Although this baseline outperforms the no-checkpoint setting, the best performance is achieved when both checkpoints are applied with uniformly distributed tokens, indicating the benefits of both the reduced complexity and the coarse-to-fine design.
Finally, we visualize the training loss ($\mathcal{L}_{tok}$) and validation performance in \cref{fig:utc_curve}. The curves demonstrate that our UTC leads to faster convergence, which is reflected by lower training loss and higher validation performance at the same epoch.

\begin{table}[t]
    \small
    \centering
    \begin{minipage}[b]{0.6\linewidth}
        \centering
        \setlength\tabcolsep{1pt}
        \resizebox{\linewidth}{!}{
        \begin{tabular}{l|l|ccccc}
        \toprule
        \multirow{2}{*}{Condition} & \multirow{2}{*}{Fusion Strategy} & \multicolumn{2}{c}{DTW-JPE$\downarrow$} & \multicolumn{2}{c}{SiBLEU$\uparrow$} & \multicolumn{1}{c}{SiCLIP$\uparrow$} \\
        \cmidrule(r){3-4}\cmidrule(r){5-6}\cmidrule(r){7-7}
        & & Body & Hands & Body & Hands & R@1 \\
        \midrule
    
        N/A & Average \cite{zuo2025soke} & 6.35 & 8.50 & 4.06 & 3.42 & 35.95 \\
        \midrule

        \multirow{6}{*}{Codebook} & Average & 6.19 & 8.25 & 4.45 & 3.73 & 36.98 \\
        & MLP & 6.22 & 8.29 & 4.19 & 3.86 & 37.02 \\
        & Transformer & 6.17 & 8.13 & 4.33 & 3.62 & 36.49 \\
        & Sparse MoE (Top-1) & 6.48 & 8.74 & 3.76 & 3.27 & 35.65 \\
        & Sparse MoE (Top-2) & 6.29 & 8.41 & 3.91 & 3.50 & 36.87 \\
        & Dense MoE (ours) & \textbf{5.99} & \textbf{7.96} & \textbf{4.87} & \textbf{4.08} & \textbf{38.61} \\
        
        \bottomrule
        \end{tabular}
        }
        \caption{Ablation study for the MoP embedding. }
        \label{tab:abl_emb}
    \end{minipage}
    \hfill
    \begin{minipage}[b]{0.39\linewidth}
        \centering
        \includegraphics[width=\linewidth]{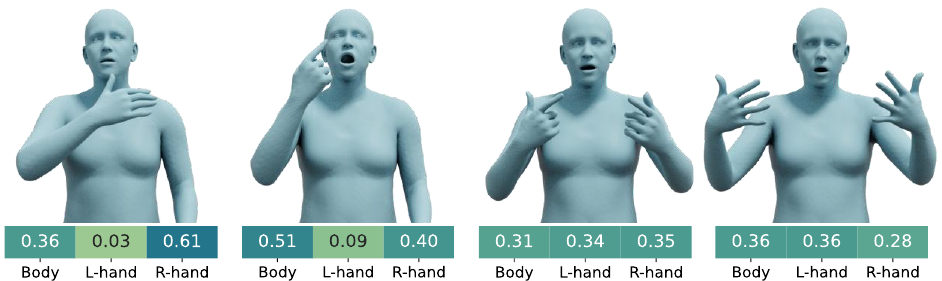}
        \captionof{figure}{Visualization of the learned gating weights $\mathbf{G}$.}
        \label{fig:vis_gate}
        \vspace{-3mm}
        \resizebox{1.0\linewidth}{!}{
        \begin{tabular}{l|c|c|c}
        \toprule
        Category & Left & Right & Two \\
        \midrule
        Weights & .41/.47/.12 & .47/.08/.45 & .32/.31/.37 \\
        \bottomrule
        \end{tabular}
        }
        \captionof{table}{Weight distribution.}
        \label{tab:mop_weight}
    \end{minipage}
    \vspace{-13mm}
\end{table}

\noindent\textbf{Mixture-of-Parts (MoP) Embedding Layer.}
Our MoP sign embedding layer dynamically fuses token embeddings from different body parts via a learnable gate. As shown in \cref{tab:abl_emb}, MoP significantly outperforms the prior design that simply averages randomly initialized part-wise token embeddings \cite{zuo2025soke}.
When conditioned on well-trained codebooks, we evaluate three standard fusion strategies: 1) averaging, 2) an MLP applied to the concatenation of part-wise code embeddings, and 3) a two-layer Transformer equipped with a $|\mathrm{CLS}|$ token. Results show that all conditioned fusion strategies outperform the non-conditioned baseline \cite{zuo2025soke}, validating the effectiveness of codebook conditioning.

Our MoP is built upon a dense mixture-of-experts (MoE) formulation \cite{cai2025moe_survey}, where all experts (corresponding to body parts) are activated and weighted by the gating network. As reported in \cref{tab:abl_emb}, we compare this design against sparse MoE variants that fuse only the body parts associated with the top-1 or top-2 gate activations. These sparse configurations consistently yield inferior performance relative to the default dense setting.

To further analyze the learned gating behavior, we visualize MoP gate weights for four example signs in \cref{fig:vis_gate}. For single-handed signs (left two examples), the MoP layer assigns higher weights to the active hand, indicating that it effectively captures hand dominance. In contrast, for two-handed signs, the layer learns comparable weights for both hands, reflecting balanced contributions during generation.
Quantitatively, we identify the signing hand using a simple rule based on wrist position relative to the spine-01 joint \cite{smplx}, and accordingly classify signs as single- or two-handed. \cref{tab:mop_weight} reports the correlation between MoP gate weights (body/left hand/right hand) and hand dominance, indicating that the learned gates align with interpretable physical structure.

\begin{wraptable}{r}{0.6\linewidth}
\small
\setlength\tabcolsep{2pt}
    \vspace{-6mm}
    \centering
    \resizebox{\linewidth}{!}{
    \begin{tabular}{l|cc|cc|cc}
    \toprule
    \multirow{2}{*}{Method} & \multicolumn{2}{c|}{CSL-Daily$\uparrow$} & \multicolumn{2}{c|}{Phoenix$\uparrow$} & \multicolumn{2}{c}{How2Sign$\uparrow$} \\
    \cmidrule(r){2-3}\cmidrule(r){4-5}\cmidrule(r){6-7}
    & R@1 & R@5 & R@1 & R@5 & R@1 & R@5 \\
    \midrule

    Baseline (CLIP) \cite{jiang2024signclip} & 41.07 & 81.97 & 58.41 & 88.01 & 9.27 & 28.47 \\
    + Decoupled encoders & 45.07 & 85.88 & 61.84 & 90.03 & 12.13 & 38.48 \\
    + Fine-grained loss & \textbf{59.10} & \textbf{95.66} & \textbf{68.85} & \textbf{95.64} & \textbf{41.33} & \textbf{88.43} \\
    
    \bottomrule
    \end{tabular}
    }
    \vspace{-3mm}
    \caption{Ablation study for the SiCLIP.}
    \label{tab:abl_clip}
    \vspace{-5mm}
\end{wraptable}

\noindent\textbf{SiCLIP.}
To better evaluate the semantic alignment between texts and generated signs, we follow established practices in related fields \cite{tevet2022motionclip,meng2025rethinking} and train a CLIP model over ground-truth sign motions and texts, referred to as SiCLIP. We assess its performance using the sign-to-text retrieval task. 
As shown in \cref{tab:abl_clip}, simply using a contrastive loss \cite{jiang2024signclip} results in low retrieval performance. Introducing decoupled sign encoders \cite{zuo2025soke} yields noticeable improvements. Most of the gains come from the fine-grained contrastive loss \cite{cheng2023cico}, which implicitly learns precise word-level alignments.
The final retrieval performance is on par with that of a leading motion retrieval model \cite{petrovich2023tmr}, validating the reliability of SiCLIP as an SLG evaluator.

\vspace{-3mm}
\section{Conclusion}
We present MaDiS, a novel SLG approach built upon MDLMs that enable bidirectional context modeling and efficient generation.
To effectively inject sign language knowledge into MDLMs, we design a tri-level cross-modal pretraining framework that jointly learns from token, latent, and 3D physical spaces.
To improve training efficiency of MDLMs, we simplify unmasking orders by proposing a novel unmasking strategy with temporal checkpoints. 
Moreover, a mixture-of-parts sign embedding layer is introduced to enhance part-wise token integration through gated, codebook-conditioned fusion.
Experiments on standard SLG benchmarks demonstrate that MaDiS achieves SOTA performance, while improving inference throughput by 40\%.

\noindent\textbf{Acknowledgements.}
S. Zafeiriou and part of the research was funded by the EPSRC Fellowship DEFORM (EP/S010203/1), EPSRC Project GNOMON (EP/X011364/1) and Turing AI Fellowship (EP/Z534699/1). R.A. Potamias and R. Zuo were supported by EPSRC Project GNOMON (EP/X011364/1). J. Deng was supported by the NVIDIA Academic Grant.
We would also like to thank Yuecong Min for coordinating the user study, and the anonymous signers who participated in it.

The authors acknowledge the use of resources provided by the Isambard-AI National AI Research Resource (AIRR). Isambard-AI \cite{mcintosh2024isambard} is operated by the University of Bristol and is funded by the UK Government’s Department for Science, Innovation and Technology (DSIT) via UK Research and Innovation; and the Science and Technology Facilities Council [ST/AIRR/I-A-I/1023].

%
%
\bibliographystyle{splncs04}
\bibliography{main}
\end{document}